\documentclass[letterpaper]{article} 
\usepackage{aaai2026}  
\usepackage{times}  
\usepackage{helvet}  
\usepackage{courier}  
\usepackage[hyphens]{url}  
\usepackage{graphicx} 
\usepackage{natbib}  
\usepackage{caption} 
\usepackage{algorithm}
\usepackage{algorithmic}
\usepackage[T1]{fontenc}

\usepackage{newfloat}
\usepackage{listings}
\DeclareCaptionStyle{ruled}{labelfont=normalfont,labelsep=colon,strut=off} 
\floatstyle{ruled}
\newfloat{listing}{tb}{lst}{}
\floatname{listing}{Listing}
\usepackage{xcolor}
\usepackage{amsmath}
\usepackage{multirow}
\usepackage{booktabs}
\usepackage{colortbl}
\usepackage{tcolorbox   }
\usepackage{array}
\nocopyright 
\definecolor{customblue}{RGB}{31, 107, 175}
\definecolor{mygray}{gray}{0.9}

\tcbset{
    mybox/.style={
        colframe=customblue,          
        colback=customblue!5,        
        coltitle=white,         
        colbacktitle=customblue,     
        fonttitle=\bfseries,    
        boxrule=0.5mm,         
        arc=3mm,             
        width=\linewidth,        
        left=1mm,                
        right=1mm,              
        top=1mm,                
        bottom=1mm,            
        fontupper=\color{black}   
    }
}

\title{Why Do Open-Source LLMs Struggle with Data Analysis?\\ A Systematic Empirical Study}

\author{
  Yuqi Zhu${^{\spadesuit\heartsuit}}$, 
  Yi Zhong$^{\spadesuit}$, 
  Jintian Zhang${^{\spadesuit\heartsuit}}$,
  \textbf{Ziheng Zhang}${^{\diamondsuit}}$,
  \textbf{Shuofei Qiao}$^{\spadesuit\heartsuit}$, \\
  \textbf{Yujie Luo}$^{\spadesuit\heartsuit}$,  
  \textbf{Lun Du}$^{\clubsuit\heartsuit}$, 
  \textbf{Da Zheng}$^{\clubsuit\heartsuit}$, 
  \textbf{Ningyu Zhang}$^{\spadesuit\heartsuit}$\footnotemark[1], 
  \textbf{Huajun Chen}$^{\spadesuit\heartsuit}$\thanks{$\quad$ Corresponding Author.}  \\
}

\affiliations{
   $^\spadesuit$Zhejiang University ~$^\clubsuit$Ant Group ~$^\diamondsuit$Independent Researcher\\ 
  $^\heartsuit$Zhejiang University - Ant Group Joint Laboratory of Knowledge Graph  \\
  \texttt{\{zhuyuqi,zhangningyu\}@zju.edu.cn} \\
}

\begin{document}
\maketitle
\begin{abstract}
Large Language Models (LLMs) hold promise in automating data analysis tasks, yet open-source models face significant limitations in these kinds of reasoning-intensive scenarios. In this work, we investigate strategies to enhance the data analysis capabilities of open-source LLMs. By curating a seed dataset of diverse, realistic scenarios, we evaluate model behavior across three core dimensions: data understanding, code generation, and strategic planning. Our analysis reveals three key findings: (1) Strategic planning quality serves as the primary determinant of model performance; (2) Interaction design and task complexity significantly influence reasoning capabilities; (3) Data quality demonstrates a greater impact than diversity in achieving optimal performance. We leverage these insights to develop a data synthesis methodology, demonstrating significant improvements in open-source LLMs' analytical reasoning capabilities.
\end{abstract}

\begin{links}
    \link{Code}{https://github.com/zjunlp/DataMind}
    \link{Extended version}{https://arxiv.org/abs/2506.19794}
\end{links}

\section{Introduction}

\begin{figure}[!t] 
    \centering
    \scalebox{1}{
    \includegraphics[width=1\linewidth]{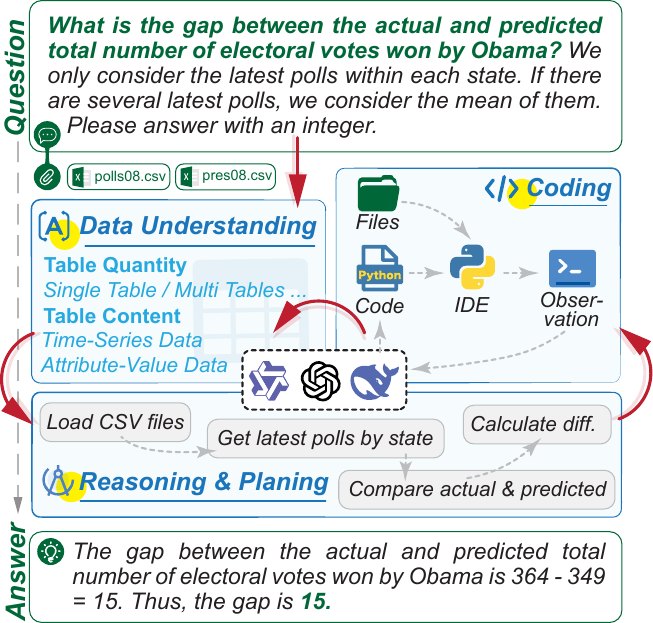} 
    }
    \caption{Core capabilities involved in data analysis tasks. We break down the process into three key components: data understanding, coding and planning.}
    \label{fig:intro}
\end{figure} 

Data analysis is a complex, interactive process central to scientific discovery, business intelligence, and decision-making~\cite{intro/donoho201750, DSInEraofAI}. 
It requires models to understand natural language queries, interpret structured data, formulate hypotheses, generate executable code, and iteratively refine reasoning—often across multiple turns of interaction (Figure~\ref{fig:intro}). 
As such, it demands tight integration of language understanding, logical reasoning, programming skills, and long-horizon planning, which together pose unique challenges beyond standard NLP tasks.

Large Language Models (LLMs) have shown promise in automating data analysis, with systems like DS-Agent~\cite{DS-Agent}, AutoML-Agent~\cite{automl-agent}, and Data Interpreter~\cite{ds/datainterpreter} demonstrating increasingly sophisticated behaviors. 
Domain-specific benchmarks such as DSBench~\cite{datasets/dsbench}, BLADE~\cite{ds/blade}, QRData~\cite{datasets/qrdata}, and DiscoveryBench~\cite{dataset/DiscoveryBench} further enable targeted evaluation. 
Yet, performance on these tasks remains dominated by large-scale, advanced models such as GPT-4~\cite{openai2024gpt4ocard} and DeepSeek-R1~\cite{deepseek-r1}, while open-source alternatives, especially smaller models, still struggle in real-world analytical settings.

This raises a key research question: \textit{\textbf{How can we effectively enhance open-source LLMs for complex, reasoning-intensive data analysis tasks?}}
Prior work has shown that fine-tuning on high-quality synthetic data can improve reasoning capabilities in domains like mathematics~\cite{s1,limo} and code generation~\cite{opencodereasoning}. 
However, for data analysis tasks that involve multi-step interactions, dynamic environments, and mixed goals, it remains unclear which properties of the training data, such as task difficulty, scenario diversity, or interaction structure, actually lead to better generalization.

In this paper, we take a capability-aware approach to understanding how to build more effective data analysis agents by analyzing the model capabilities involved in the analytical process. 
We decompose the task into three core dimensions: \textbf{Data Comprehension}, \textbf{Code Generation}, and \textbf{Strategic Planning}.
Leveraging a curated seed dataset that encompasses diverse data analysis scenarios, we conduct targeted experiments and ablation studies to analyze the factors influencing model generalization and performance on complex analytical tasks. 
Here, our analysis reveals three key findings: 
\begin{itemize} 
\item The model's planning ability emerges as a more critical determinant of success than its capabilities in data understanding and code generation. This underscores the importance of strategic foresight and structured reasoning in navigating complex analytical scenarios.
\item Appropriate interaction turns, combined with the complexity of the data and suitable reasoning descriptions, can enhance the model's reasoning capacity.
However, performance gains vary across different data analysis tasks, suggesting that task-specific characteristics play a crucial role in shaping the model's effectiveness.
\item High-quality training data proves to be more critical than data diversity for achieving optimal performance in data analysis tasks. 
This emphasizes the necessity of curating datasets with precise and comprehensive annotations to ensure reliable and robust model outcomes.
\end{itemize}

To substantiate these insights, we propose a strategy-guided data synthesis framework that leverages empirical findings to inform key design choices, such as selecting informative interaction patterns and enriching data with concise reasoning traces, to enable more effective model learning. 
By fine-tuning on the resulting dataset, we demonstrate measurable improvements in performance, achieving results competitive with leading closed-source models.

\section{Background of Data Analysis Agents}
Data analysis tasks aim to derive actionable insights from data through systematic exploration and analysis~\cite{survey/DA_IN_GenAI,survey/LLMforDS}. 
In a typical workflow, analysts begin with specific questions about a dataset, then proceed through multiple analytical steps. 
These steps include data preprocessing and cleaning, hypothesis exploration, data transformations, and report generation.
The process is inherently interactive - analysts work with structured tabular data, develop analysis code, interpret intermediate results, and iteratively refine their analysis before generating final reports with their findings. 

To formally characterize the data analysis process for LLM agents, we define a parameterized analysis function \(f_\theta\) that maps the input components to the analytical outputs:
\begin{align}
f_\theta: (\mathcal{D}, \mathcal{Q}, \mathcal{T}) \rightarrow (\mathcal{S}, \mathcal{R})
\end{align}
where \(\mathcal{D}\) represents the structured data, \(\mathcal{Q}\) specifies the analytical objective or query, and \(\mathcal{T}\) is the library of available analysis tools. 
\(f_\theta\) models the behavior of an LLM-based agent that performs analysis by generating a sequence of intermediate analysis states \(\mathcal{S}=\{s_t\}\), ultimately producing a final report \(\mathcal{R}\) to summarize the results.

\section{Experimental Settings}
\paragraph{Evaluation Protocol}
We adopt a capability-aware evaluation protocol aligned with our decomposition into Data Comprehension, Code Generation, and Strategic Planning. 
For the former two, we use \textit{prompt-based evaluation}; for Strategic Planning, we employ \textit{LoRA-based fine-tuning}~\cite{lora} to target long-horizon reasoning. 
\paragraph{Model Selection}
We evaluate a series of open-source model variants, including Qwen2.5-7B-Instruct, Qwen2.5-7B-Coder, Qwen2.5-7B-Instruct-1M, and R1-Distill-Qwen-7B, alongside strong API-based baselines like GPT-4o, DeepSeek-v3\footnote{\texttt{gpt-4o-2024-08-06}, \texttt{deepseek-v3-0324}}, and DeepSeek-R1. 
All models follow the ReAct framework~\citep{iclr23/react} for multi-turn interaction, iteratively alternating between planning, code generation, and execution.

\paragraph{Data Collection and Curation}
To construct the training dataset, we collect samples from DAEval~\cite{datasets/icml/daeval} (real-world CSVs from GitHub), DSBench~\cite{datasets/dsbench} (ModelOff competition tasks), TableBench~\cite{datasets/aaai/tablebench} (multi-domain tabular reasoning), WTQ~\cite{datasets/acl/wtq}, and FetaQA~\cite{datasets/tacl/fetaqa}. 
To enrich reasoning diversity, we generate additional synthetic samples all using DeepSeek-R1.

All training data are collected to ensure no overlap with our evaluation benchmarks, enabling a reliable out-of-distribution (OOD) assessment.
Utilizing correctness-based filtering, we retain \textbf{6,443} distinct samples, encompassing a diverse array of analytical challenges.
To ensure data quality, we apply a two-stage filtering process. 
First, we remove invalid samples based on several criteria: low-quality code implementations, such as those failing to utilize provided files or producing code with no meaningful return values; samples containing compilation errors; and entries that do not conform to format requirements. 
Following the automated filtering, we perform manual verification through sampling to further refine the dataset. 
The final corpus contains \textbf{5,613} high-quality instances used for fine-tuning.

\paragraph{Evaluation}
We evaluate on two comprehensive benchmarks: \textbf{DiscoveryBench}~\cite{dataset/DiscoveryBench}, which includes 239 real-world tasks from its 264-task suite across six domains (e.g., sociology, engineering), and \textbf{QRData}~\citep{datasets/qrdata}, a benchmark designed for statistical and causal analysis comprising 411 questions paired with data sheets from textbooks, academic papers, and online resources.

Following~\citet{ds/Reproducibility}, we use accuracy as the evaluation metric. 
Given that both predictions and references are in natural language, we employ GPT-4o-mini\footnote{\texttt{gpt-4o-mini-2024-07-18}} for agreement scoring.
Further details on dataset statistics and training configurations are provided in Appendix ~\ref{app:experiment_setup}.

\section{Core Capabilities for Data Analysis}

Data analysis tasks present a unique challenge, requiring the integration of multiple capabilities.
We identify three core skills that determine model performance: \textbf{Data Comprehension}, the ability to understand and effectively utilize structured data; \textbf{Code Generation}, the skill of producing correct and efficient analytical code; and \textbf{Strategic Planning}, the capacity to decompose complex problems into manageable steps. 
All ablation experiments use comparable dataset sizes (via subsampling) to ensure fair within-condition comparisons. 
Slight variations across modules reflect differing experimental goals.
This setup supports a systematic evaluation of the data characteristics that underlie effective analytical reasoning, thereby laying the foundation for our investigation.

\subsection{Data Comprehension}
\label{sec:data_comprehension}
To investigate whether data comprehension serves as a critical factor in enabling effective data analysis, we design a series of experiments to evaluate the model's ability to reason over structured information.
Specifically, the experiments focus on two key aspects: (1) whether explicitly providing structured context, such as tabular data, enhances the model's reasoning accuracy, and (2) how the model performs when faced with increased complexity, including scenarios involving multiple sources of structured information, some of which may be irrelevant or distractive.

\paragraph{Tabular Information.}
To evaluate the impact of tabular data visibility, we compare two settings: without (\textbf{w/o Info}) and with (\textbf{w/ Info}) table information. 
In the \textbf{w/ Info} setting, we provide the necessary table information, such as column names, data types, and sample entries; in \textbf{w/o Info}, only the filename (e.g., \texttt{data.csv}) is provided. 
This ablation tests whether explicit access to table context improves reasoning.
As shown in Table~\ref{tab:with_tableinfo}, adding table information helps the models in simpler tasks like QRData, showing slight improvements in performance.
However, the gains are limited, indicating that the models already handle much of the reasoning without explicit table inputs.
For the more complex DiscoveryBench, while the 7B model benefits from the inclusion of table information, the 14B model exhibits a slight drop in accuracy. 
This decline may be related to the increased input length, which could lead the 14B model to generate longer, less focused outputs. 
We hypothesize that this affects reasoning coherence, though further analysis is needed to confirm the effect.

\begin{table}[!t]
\centering
\small 
\scalebox{0.9}{ 
\begin{tabular}{>{\centering\arraybackslash}p{1.8cm}|>{\centering\arraybackslash}p{1.3cm}|>{\centering\arraybackslash}p{1.3cm}|>{\centering\arraybackslash}p{1.3cm}|>{\centering\arraybackslash}p{1.3cm}}
\toprule
\multirow{2}{*}{\textbf{Model}} & \multicolumn{2}{c|}{\textbf{QRData}} & \multicolumn{2}{c}{\textbf{DiscoveryBench}} \\
\cmidrule(lr){2-5}
 & \textbf{w/o Info} & \textbf{w Info} & \textbf{w/o Info} & \textbf{w Info} \\
\midrule
Qwen2.5-7B & 6.57 &7.54 & 0.42 & 1.26 \\  
Qwen2.5-14B &15.09 & 15.82 & 0.42 & 0.00 \\  
\bottomrule
\end{tabular}
}
\caption{Accuracy comparison of 7B and 14B models on QRData and DiscoveryBench (w/o and w/ table information).}
\label{tab:with_tableinfo}
\end{table}

\paragraph{Data Complexity.}
To evaluate the model's ability to maintain focus amid distracting information, we introduce additional tables as irrelevant inputs, which act as semantic noise.
This setup requires the model to reason over multiple tables, some of which are irrelevant, simulating scenarios with heightened complexity. 
Models that effectively filter out irrelevant data while focusing on task-relevant information are considered to exhibit stronger data understanding capabilities.
For a fair evaluation, we remove task-specific background descriptions from the input, forcing the model to rely solely on the tabular inputs alone.
As shown in Table~\ref{tab:model_accuracy_comparison}, the inclusion of redundant data increases input complexity but does not lead to a significant decline in overall performance. 
The 7B model exhibits a modest decline, suggesting it is more sensitive to increased input noise.
In contrast, the 14B model maintains stable performance, demonstrating stronger filtering capability and resilience in multi-source settings.

\begin{table}[!t]
\centering
\small 
\scalebox{0.9}{ 
\begin{tabular}{>{\centering\arraybackslash}p{1.8cm}|>{\centering\arraybackslash}p{1.4cm}|>{\centering\arraybackslash}p{1.2cm}|>{\centering\arraybackslash}p{1.4cm}|>{\centering\arraybackslash}p{1.2cm}}
\toprule
\multirow{2}{*}{\textbf{Model}} & \multicolumn{2}{c|}{\textbf{QRData}} & \multicolumn{2}{c}{\textbf{DiscoveryBench}} \\
\cmidrule(lr){2-5}
& \textbf{w/o Extra} & \textbf{w/ Extra} & \textbf{w/o Extra} & \textbf{w/ Extra}\\
\midrule
Qwen2.5-7B &37.96 &34.55 &5.44 &4.18 \\  
Qwen2.5-14B &52.55 &52.07 &10.88 &12.13 \\  
\bottomrule
\end{tabular}
}
\caption{Accuracy comparison of 7B and 14B models on QRData and DiscoveryBench (w/o and w/ extra data files).}
\label{tab:model_accuracy_comparison}
\end{table}

\paragraph{Discussion and Implications.}
The minimal performance gains from explicit table input suggest that data comprehension is not the primary bottleneck in data analysis. 
Even under input noise, models maintain stable performance and demonstrate robust data comprehension abilities—likely internalized during pretraining.
This indicates that basic data understanding has already been internalized during pretraining, rendering additional context less beneficial. 


\begin{table}[!t]
\centering
\small
\scalebox{1}{
\begin{tabular}{lcc}
\toprule
\textbf{Model} & \textbf{QRData} & \textbf{DiscoveryBench} \\ 
\midrule
\rowcolor{mygray}
 
\rowcolor{mygray}
\multicolumn{3}{l}{\textbf{Multi-Turn Setting}} \\ 
\midrule
 Qwen2.5-7B-Instruct  &   39.71 &   14.64 \\ 
 Qwen2.5-14B-Instruct &   53.53 &  24.27 \\
 Qwen2.5-32B-Instruct &  57.18 &  27.62 \\ \midrule
 Qwen2.5-7B-Coder        & 36.50 & 13.60 \\
Qwen2.5-7B-Instruct-1M   & 39.17 & 15.48  \\
 R1-Distill-Qwen-7B       & 30.41 & 7.95  \\  \midrule
 GPT-4o & 59.85  & 28.03  \\ 
 DeepSeek-v3 & 65.21   &  36.82 \\ 
 Deepseek-R1&  63.26   &  37.66   \\ 

\bottomrule
\end{tabular}
}
\caption{Performance comparison across models in multi-turn settings (\% accuracy).}
\label{table:baselines}
\end{table}

\subsection{Code Capability}
\label{sec:code_capability}
To investigate the role of code generation in data analysis, we evaluate a diverse set of models with varying training objectives and architectures. 
Rather than treating code correctness as an end in itself, we examine how well models utilize code as part of a broader reasoning process to achieve task success.

\paragraph{Code Performance.}
We analyze how different models utilize code during problem solving. 
Table~\ref{table:baselines} provides the overall performance across models, while Table~\ref{table:error_code} summarizes the average code error rates observed across two datasets.

Our analysis reveals several key findings: (1) \textbf{Code specialization does not guarantee better performance}: Qwen2.5-7B-Coder does not demonstrate a clear advantage over general-purpose models. 
This suggests that code specialization alone may not translate to better performance in analytical tasks, due to limitations in instruction-following or reasoning generalization.
(2) \textbf{Distillation can lead to functional hallucination}: R1-Distill-Qwen-7B, despite being distilled from a large reasoning model, performs poorly, often hallucinating file interpretations rather than generating executable code.
(3) \textbf{Long-context capability does not imply efficient task execution}: When comparing Qwen2.5-7B-Instruct-1M and Qwen2.5-7B-Instruct under matched output length constraints, both models exhibit comparable coding capabilities; however, the latter demonstrated superior planning efficiency by completing tasks in fewer interaction rounds.

\begin{figure}[!t] 
    \centering
    \scalebox{0.9}{
    \includegraphics[width=1\linewidth]{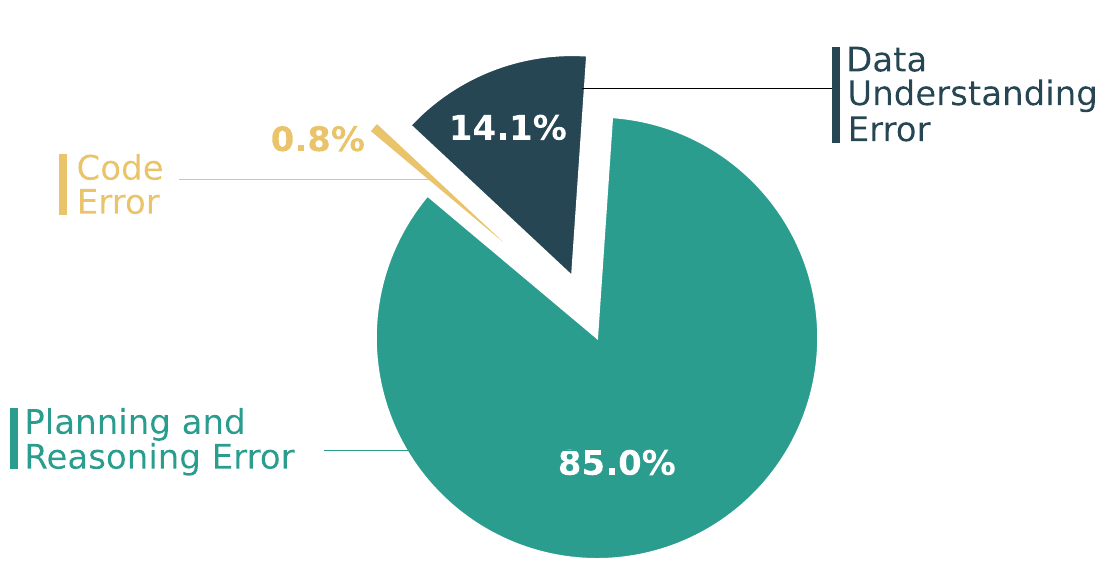} 
    }
    \caption{The distribution of error type.}
    \label{fig:exp_error}
\end{figure}

To further substantiate these findings, we manually sample 354 erroneous responses and categorized the errors using GPT-4o-mini. 
The categorization is based on the gap between the incorrect responses and the corresponding correct trajectories. 
As shown in Figure~\ref{fig:exp_error}, only a small fraction of errors stem from syntactic or semantic code defects (e.g., invalid syntax). 
The majority of errors stem from higher-level reasoning failures, such as incorrect hypothesis formulation or premature termination. This indicates that \textit{planning}, rather than \textit{coding}, is more important.

\begin{table}[!t]
\centering
\small
\scalebox{0.9}{
\begin{tabular}{lcc}
\toprule
\textbf{Model} & \textbf{QRData} & \textbf{DiscoveryBench} \\ 
\midrule
 Qwen2.5-7B-Instruct  &   34.64\% &   54.25\% \\ 
 Qwen2.5-14B-Instruct &   29.94\% &  40.64\% \\
 Qwen2.5-32B-Instruct &  20.73\% &  31.73\% \\ \midrule
 Qwen2.5-7B-Coder        & 43.30\% & 50.98\% \\
Qwen2.5-7B-Instruct-1M   & 31.44\% & 55.00\%  \\
 R1-Distill-Qwen-7B       & 48.37\% & 60.00\%  \\  
\bottomrule
\end{tabular}
}
\caption{Average code error rate of different models.}
\label{table:error_code}
\end{table}

\paragraph{Discussion and Implications.} 
Our results indicate that while coding proficiency is necessary, it may not be the primary determinant of success in data analysis. 
Modern instruction-tuned models already possess sufficient code generation capabilities to handle typical analytical operations. 
What truly distinguishes stronger agents may be their ability to \textit{strategically deploy code}—to select appropriate operations, sequence them logically, and interpret outputs for iterative reasoning.

\subsection{Strategic Planning}
\label{sec:strategic_planning}

Data analysis is inherently a planning-intensive process that demands careful coordination of data access, transformation, and reasoning steps. 
Building on our earlier findings that strategic reasoning plays a decisive role in task success, we further investigate how the reasoning capabilities of LLMs shape their performance in complex analytical workflows.
Given that instruction-tuned models (e.g., Qwen2.5-7B-Instruct) demonstrate more coherent planning behavior, we use them as the foundation for all subsequent experiments. 
Specifically, we systematically evaluate model performance along four key aspects: Interaction Turns, Reasoning Length, Task Complexity, and Problem Diversity.

\subsubsection{Interaction Turns.}
To assess the impact of dialogue turn strategies on model performance, we categorized interactions into three primary turn lengths: \textbf{Short} (2-3 turns), \textbf{Medium} (4-5 turns), and \textbf{Long} (6+ turns).
Additionally, we included a \textbf{Mixed} strategy that combines varying turn lengths to reflect more dynamic interaction scenarios.

\begin{figure}[!t] 
    \centering
    \scalebox{0.9}{
    \includegraphics[width=1\linewidth]{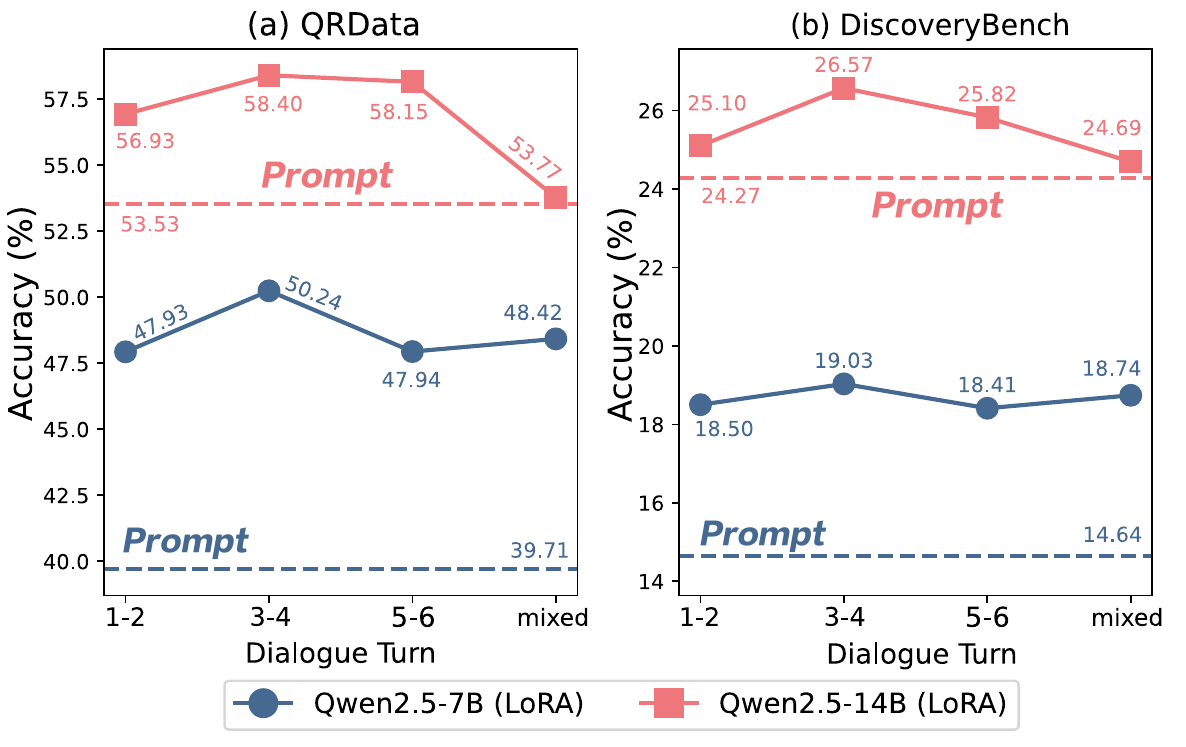} 
    }
    \caption{Impact of dialogue turn strategies across different Qwen model scales and training methods.}
    \label{fig:exp_turn_len_7_14b}
\end{figure}

Figure~\ref{fig:exp_turn_len_7_14b} reports the performance of Qwen2.5-7B and Qwen2.5-14B under different dialogue strategies. 
To ensure a fair comparison, both models were fine-tuned using the same dataset size across all strategies, which is 1020 here.
The results reveal consistent trends across the two models: medium-length interactions generally achieve relatively better performance across both datasets, suggesting they provide an optimal balance between reasoning depth and focus.
In contrast, short and long interaction strategies yield slightly lower yet relatively stable results. 
Notably, the mixed strategy consistently underperforms, likely because variable turn lengths disrupt the model’s ability to learn stable interaction patterns.
Given the alignment in trends across model scales and the computational efficiency of the 7B variant, we select Qwen2.5-7B for all subsequent experiments. This enables deeper ablation studies while maintaining analytical rigor.

\begin{table}[!t]
\centering
\small 
\scalebox{0.9}{
\begin{tabular}{l|c|cc}
\toprule
\textbf{Turn Category} 
& \textbf{\# Sample}
& \textbf{QRData} 
& \textbf{DiscoveryBench} \\
\midrule
\textit{All}  & 5613  & 48.66  & 15.00  \\  
\midrule
\textit{Short}  & 1034  & 47.68  & 23.85  \\
\textit{Medium}   & 3559 & 49.15 & 18.83  \\
\textit{Long}   & 1020 & 47.94 & 18.41  \\
\midrule
\textit{Medium + Short} & 4593 & 47.45  & 21.34  \\
\textit{Medium + Long} & 4579  & 46.96   & 21.76 \\
\bottomrule
\end{tabular}
}
\caption{Performance across different training data turn lengths (\% accuracy).}
\label{tab:turn_length_analysis}
\end{table}

To better understand the impact of turn length, we examine the performance of various interaction strategies using the full collected dataset.
The results are presented in Table~\ref{tab:turn_length_analysis}, from which we derive the following observations:
\paragraph{\textit{(i)} Turn length preferences are task-dependent.}
The results reveal that  Medium-length turns achieve a relatively higher performance on QRData, indicating their suitability for tasks requiring moderate reasoning depth. 
On DiscoveryBench, which features longer and more complex inputs, Short turns surprisingly outperform other strategies, potentially due to their ability to focus on concise and straightforward reasoning. 
\paragraph{\textit{(ii)} Data quality outweighs quantity.}
Our experiments reveal that increasing the amount of training data does not necessarily lead to better performance, even when using the same interaction turn strategies.
In fact, medium-length turns trained on a smaller subset consistently outperform models trained on the full dataset.
This finding highlights the importance of data quality and task relevance, suggesting that the effectiveness of fine-tuning is shaped by factors beyond data quantity alone.
\paragraph{\textit{(iii)} Mixed strategies require principled design.} 
While combining turn lengths may seem beneficial for diversity, the underperformance of mixed strategies suggests that unstructured variation introduces inconsistency, hindering the model's ability to learn coherent reasoning patterns. 
Rather than improving flexibility, random mixing may confuse policy learning. 
Effective use of diverse interaction styles likely requires intentional design, such as curriculum scheduling or adaptive control, rather than uniform combination.

To ensure consistency and facilitate direct comparisons in subsequent experiments, we adopt the \textbf{Medium} turn strategy as the baseline for all follow-up evaluations.



\subsubsection{Reasoning Length.}
To investigate whether longer reasoning chains from stronger models improve planning and task success, we augment training samples with intermediate \texttt{<think>} segments generated by DeepSeek-R1. 
These segments aim to capture intermediate reasoning steps that may scaffold better decision-making during multistep analysis.

\begin{figure}[!t] 
    \centering
    \scalebox{1}{
    \includegraphics[width=1\linewidth]{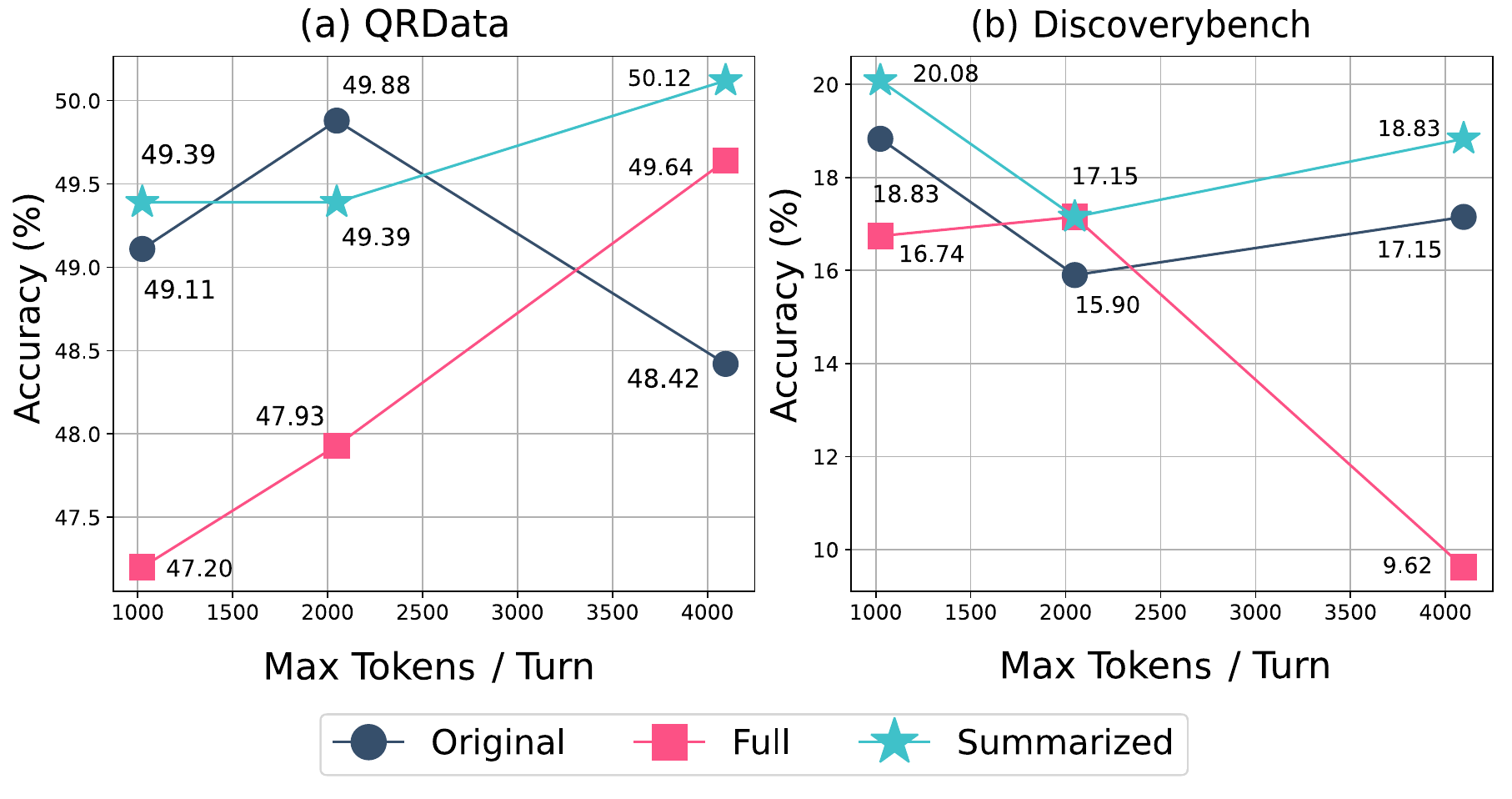} 
    }
    \caption{Impact of reasoning length on model performance across token budgets.}
    \label{fig:exp_performance_comparison}
\end{figure}

We evaluate three settings: (1) \textbf{Original reasoning}, which uses the original training data without modification. These samples include reasoning content, but the reasoning is typically shorter and less explicit compared to the augmented settings; (2) \textbf{Full reasoning}, which replaces the original reasoning with full \texttt{<think>} traces for intermediate turns (excluding the first and final turns); and (3) \textbf{Summarized reasoning}, which substitutes the original reasoning with concise summaries (generated by an LLM) of the full traces. 
The reasoning is inserted only in the middle turns to avoid hallucinations during initial grounding and to maintain brevity in the final answer generation. 
Experiments are conducted under varying per-turn token budgets (1024, 2048, and 4096) to simulate different interaction constraints.
Figure~\ref{fig:exp_performance_comparison} presents the results. We observe that:

\paragraph{\textit{(i)} Longer reasoning is not always better.}
While longer reasoning chains might seem beneficial, the \textbf{Full} setting consistently underperforms the \textbf{Original} across most configurations, particularly in DiscoveryBench.
The sharp decline in performance at the 4096-token level suggests that excessive, unfiltered reasoning can overwhelm the model, either by introducing redundancy, amplifying internal noise, or disrupting attention coherence. 
In contrast, the \textbf{Summarized} setting (though shorter) consistently matches or surpasses baseline accuracy.
By distilling critical inferences and removing irrelevant reasoning steps, it reduces noise and enhances focus, leading to more reliable decision-making.
These results indicate that reasoning effectiveness depends less on length and more on information relevance and logical coherence. 
Well-structured, goal-aligned reasoning often outperforms longer but unfocused alternatives.

\paragraph{\textit{(ii)} Token budgets exhibit diminishing returns.}
Increasing the per-turn token budget can improve performance by enabling better integration of reasoning and task-relevant content, as seen in tasks like QRData.
However, these gains are often limited, and in some cases, such as with DiscoveryBench, a larger token budget may even reduce performance by amplifying noise or irrelevant information.
This suggests that effectiveness in reasoning-intensive tasks depends less on context size and more on information density and coherence. 
Simply allocating more tokens without improving content quality offers limited value.


\begin{table}[!t]
\centering
\small 
\begin{tabular}{l|c|c}
\toprule
\multirow{1}{*}{\textbf{Difficulty}}  & \textbf{QRData} & \textbf{DiscoveryBench} \\
\midrule
\textit{Easy}  & 42.58 & 20.50  \\ 
\textit{Medium}   & 51.34 & 18.83 \\ 
\textit{Hard}  &  48.18 & 19.50  \\ 
\midrule
\textit{Medium} + \textit{Hard}   & 51.34  & 23.01   \\ 
\bottomrule
\end{tabular}
\caption{Performance comparison across different task complexity (\% accuracy). 
}
\label{tab:table_complexity}
\end{table}

\subsubsection{Task Complexity.}
To assess how data difficulty affects model reasoning, we classify each example based on the performance of models with varying capacities. 
Specifically, a task is labeled \textit{easy} if it can be correctly solved by Qwen2.5-7B, \textit{medium} if only Qwen2.5-14B can solve it, and \textit{hard} if it requires DeepSeek-R1 to provide the correct answer. 
Here, each level contains 733 samples to ensure fair comparison.

We train models using data of varying difficulty levels, with the results summarized in Table~\ref{tab:table_complexity}. 
On QRData, performance generally improves with training data difficulty, with the \textit{medium} setting yielding the best results. 
Combining \textit{medium} and \textit{hard} data achieves comparable or better performance across both datasets, indicating that exposure to more complex tasks enhances model generalization on structured analytical problems.
 
\begin{figure}[!t] 
    \centering
    \scalebox{1}{
    \includegraphics[width=1\linewidth]{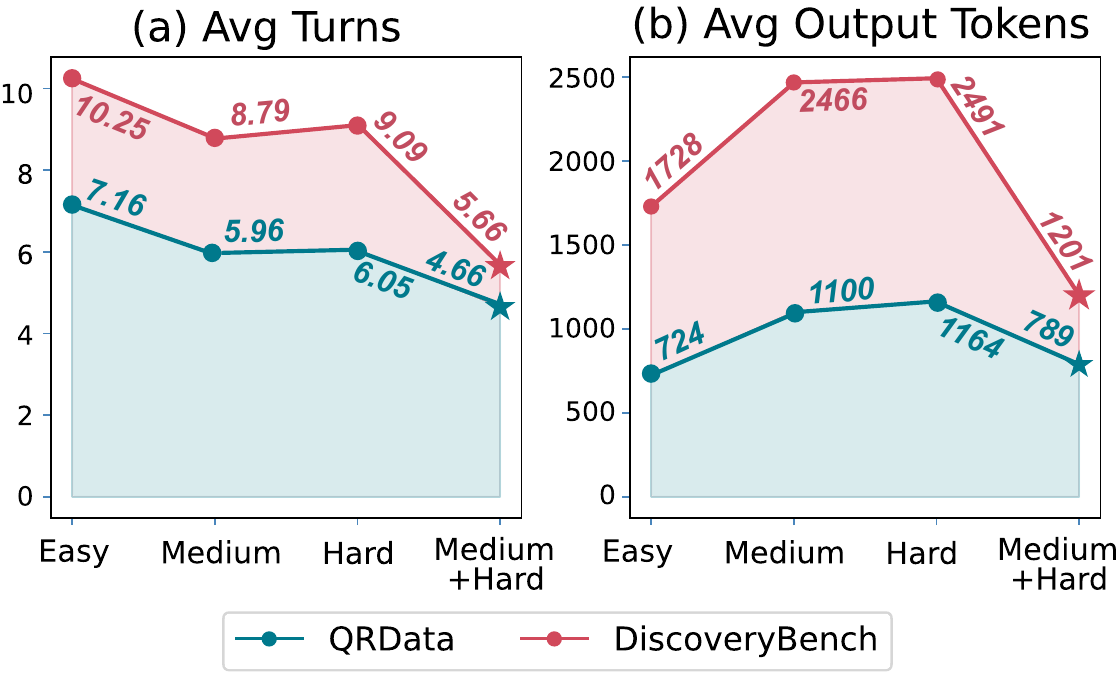} 
    }
    \caption{Impact of training data difficulty on interaction patterns. (a) Average number of response rounds of the model. (b) Average output token length of the model.}
    \label{fig:exp_complexity}
\end{figure}

To understand this effect, we analyze the models' dialogue turns and average response length, as illustrated in Figure~\ref{fig:exp_complexity}.
The results reveal that:  
\textbf{As training data difficulty increases, models shift from multi-turn, feedback-dependent interactions to generating comprehensive answers in fewer steps.}
This suggests that exposure to harder tasks encourages models to internalize reasoning steps—reducing reliance on iterative refinement and increasing reasoning density per turn. 
The result is more efficient, self-contained decision-making.

Notably, training on a mix of \textit{medium} and \textit{hard} data yields the most compact behavior, with the lowest average interaction turns and shortest responses. 
However, on DiscoveryBench, models trained on \textit{easy} data perform better. 
We attribute this to the need for fine-grained, multi-step computation, where a smaller per-step workload improves reliability.
These findings highlight a key trade-off: while complex training fosters reasoning efficiency, simpler strategies may remain valuable for tasks requiring extended exploration.

\subsubsection{Problem Diversity.}

We further investigate whether adjusting the diversity of question improves model performance. 
To this end, we annotate the dataset with semantic category labels using GPT-4o-mini and retain ten major domains after manual consolidation.

To quantify diversity, we employed a three-stage procedure to classify each question into distinct domains. 
The detailed methods and descriptions of the categories can be found in Appendix~\ref{app:problem_diverse}.
From the complete medium turn dataset, we select 2,220 examples under two settings: (1) a natural distribution that reflects the original domain frequencies, and (2) a balanced distribution that down-samples dominant domains while retaining all examples from underrepresented ones. 
Both settings preserve the overall difficulty distribution of the dataset.
Results in Table~\ref{tab:table_diversity} show minimal performance differences between the two settings, indicating that domain diversity alone does not significantly influence model performance in this context.
These findings suggest that the effectiveness of a model is not solely determined by the diversity of problems it encounters during training. 
Rather, the diversity and richness of reasoning strategies, such as the depth of reasoning processes and the complexity of logical steps, appear to play a more influential role in shaping a model's performance.

\paragraph{What Makes Data Effective ?}
The efficacy of large language models in data analysis is fundamentally shaped by the quality of the data rather than its sheer diversity. High-quality, well-structured tasks that emphasize nuanced complexity and transparent reasoning pathways are instrumental in refining the models' analytical capabilities. 

\begin{table}[!t]
\centering
\small 
\begin{tabular}{l|c|c}
\toprule
\multirow{1}{*}{\textbf{Diversity}}  & \textbf{QRData} & \textbf{DiscoveryBench} \\
\midrule
Original Distribution & 46.72 & 20.92 \\
Balanced Sampling & 45.00 & 21.76 \\

\bottomrule
\end{tabular}
\caption{Performance under different sampling strategies (\% accuracy). 
}
\label{tab:table_diversity}
\end{table}

\section{Towards Effective Data Analysis Performance}
While the previous section examined how different aspects of data analysis influence model performance, we now ask a practical question: how can these insights be leveraged to guide data collection and reuse in order to improve data analysis capabilities?

\subsection{Strategy-Guided Data Synthesis}
We design our data synthesis process in three systematically organized stages to construct a refined dataset that supports reasoning in data analysis.
1) \textbf{Prompt-Based Answer Generation.}
We begin by leveraging prompt-based generation techniques to produce multiple candidate answers for each input query. This step ensures a diverse pool of responses, capturing a range of reasoning patterns and perspectives.
2) \textbf{Targeted Instance Selection.}
Next, we prioritize medium-length dialogues and examples of medium to high difficulty, as these have been shown to facilitate more stable and effective learning. Through this filtering process, we select instances that strike a balance between complexity and informativeness.
3) \textbf{Reasoning-Driven Data Enrichment.}
Finally, we enrich each selected instance with a concise reasoning summary. This step enhances abstraction and generalization by explicitly capturing the underlying reasoning process, allowing models to better learn transferable insights.
By following this three-stage synthesis process with format standardization, we construct a dataset comprising 2.8k instances.
The resulting dataset serves as the foundation for  Supervised Fine-Tuning (SFT) to optimize model performance.
A detailed overview of the synthesis process and the training parameters is provided in Appendix~\ref{app:data_sft}.

\begin{table}[!t]
\centering
\small
\scalebox{1}{
\begin{tabular}{lcc}
\toprule
\textbf{Model} & \textbf{QRData} & \textbf{DiscoveryBench} \\ 
\midrule
\rowcolor{mygray}
\multicolumn{3}{l}{\textbf{API Models}} \\ 
\midrule
GPT-4o & 59.85  & 28.03  \\ 
DeepSeek-v3 & 65.21   &  36.82 \\ 
Deepseek-R1&  63.26   &  37.66   \\ 
\midrule
\rowcolor{mygray}
\multicolumn{3}{l}{\textbf{7B Models}} \\ 
\midrule
Qwen2.5-Instruct & 39.71 & 14.64 \\ 
Ours & 53.77  & 22.59 \\
\midrule
\rowcolor{mygray}
\multicolumn{3}{l}{\textbf{14B Models}} \\ 
\midrule
Qwen2.5-Instruct & 53.53 & 24.27 \\
Ours &  58.15 & 36.82  \\
\bottomrule
\end{tabular}
}
\caption{Performance comparison of our method and baselines across model types and sizes (\% accuracy).}
\label{table:model_api_size_comparison}
\end{table}

\subsection{Evaluation Results}

Table~\ref{table:model_api_size_comparison} presents the evaluation results of our approach, where we fine-tune 7B and 14B models on the curated dataset. 
Notably, the fine-tuned 7B model demonstrates substantial performance improvements compared to its baseline, while the fine-tuned 14B model achieves results that are comparable to or surpass those of GPT-4o.
These results suggest that even simple, insight-driven adjustments to training data can yield substantial improvements in models' performance on complex analytical tasks.
However, the performance gains appear to diminish as model scale increases, suggesting a potential saturation point. 
One possible explanation is that our strategy is constructed using Qwen2.5-7B, making the resulting training distribution better aligned with its inductive biases.
While this alignment benefits smaller architectures, it may be less effective for larger models with more diverse representational capacities.
And a key limitation of our approach lies in the dataset itself. 
While the curated dataset provides value, it falls short in addressing the diversity and complexity required for more challenging tasks, making high-quality data a persistent bottleneck.
To overcome this, future efforts should focus on expanding the dataset to include richer, more representative samples from real-world applications.
Such improvements would not only better capture the variability and nuances of real-world scenarios but also enable further refinement of filtering strategies, enhancing scalability and generalizability across different model sizes and domains.

\section{Related Work}

\paragraph{LLM Agents.}
To adapt LLMs to complex reasoning tasks, recent work has explored three main approaches: prompt engineering, supervised fine-tuning (SFT), and reinforcement learning (RL) \cite{wang2024survey,xi2025rise,guo2024large,wang2024gui,hu2025agents,wang2025large}. 
Prompt-based methods~\cite{aaai24_expel,iclr23/react, plan-and-solve} improve reasoning performance by reformulating inputs to better elicit the model’s latent capabilities. 
SFT-based methods \citep{fireact,agenttuning,lumos,autoact,knowagent} provide task adaptation through labeled data, and have been applied in settings such as S1~\cite{s1}, which introduces "budget forcing" to control reasoning steps, and LIMO~\cite{limo}, which demonstrates that complex mathematical abilities can emerge from carefully curated examples.
RL-based approaches further align model behavior with human preferences and problem-solving strategies.
Reinforcement learning further enhances reasoning capabilities by optimizing multi-step decision-making.
For example, DeepSeek-R1~\cite{deepseek-r1} leverages Group Relative Policy Optimization (GRPO) to train LLMs on complex reasoning tasks.
RAGEN~\cite{rl/ragen} introduces a modular RL framework that supports stable multi-turn learning via the StarPo architecture. 
In addition, ReSearch~\cite{rl/research} integrates RL with retrieval-augmented generation (RAG), treating search operations as part of the reasoning process and learning when and how to retrieve external information.

\paragraph{LLM agents for Data Analysis.}
Data science is an interdisciplinary field that focuses on extracting valuable insights from various data sources, with data analysis serving as a crucial component \cite{zhang2025datascibench,luo2025assistedds,gupta2025bi,sun2025agenticdata}. 
Recently, researchers have proposed several specialized LLMs to enhance automated data analysis capabilities~\cite{survey/LLMforDS,li2025autosdt}.
Data Copilot~\cite{ds/datacopilot} is a code-centric data analysis agent that efficiently handles massive data processing and visualization through pre-designed interfaces, specifically tailored to automate financial data analysis tasks while reducing errors and improving efficiency.
Data Interpreter~\cite{ds/datainterpreter} takes a different path by using hierarchical dependency graphs to represent workflows, enabling automatic task breakdown and code improvements. 
AutoKaggle~\cite{ds/autokaggle} coordinates multiple specialized agents (Planner, Developer, and Reviewer) to handle end-to-end data analysis tasks, while keeping room for human input when needed.
To evaluate the ability of these agents, researchers have developed several comprehensive benchmarks, such as InfiAgent-DABench~\cite{datasets/icml/daeval}, DSBench~\cite{datasets/dsbench}, DA-code~\cite{ds/emnlp/dacode},  and DataSciBench~\cite{ds/DataSciBench} for assessing the effectiveness of LLM-based data analysis solutions.

\section{Conclusion and Future Work}
In this work, we conduct a detailed investigation into the data efficiency challenges faced by open-source LLMs in data analysis tasks. 
By curating a dataset specifically for data analysis scenarios, we systematically assess how data structure, interaction patterns, and training strategies influence model performance. 
Our findings highlight that careful multi-turn data design and appropriately structured training data are critical for enhancing LLM reasoning in data analysis.
In future work, we will focus on expanding our dataset and exploring broader model evaluations to further advance LLMs for data analysis.

\section*{Acknowledgement}
We would like to express our sincere gratitude to the anonymous reviewers for their thoughtful and constructive feedback. This work was supported by the National Natural Science Foundation of China (No. 62206246, No. NSFCU23B2055, No. NSFCU19B2027), the Fundamental Research Funds for the Central Universities (226-2023- 00138), Ningbo Natural Science Foundation (2024J020), Yongjiang Talent Introduction Programme (2021A-156-G), and Information Technology Center and State Key Lab of CAD\&CG, Zhejiang University. This work was supported by Ant Group and Zhejiang University - Ant Group Joint Laboratory of Knowledge Graph.

\bibliography{custom}

\appendix

\section{Experimental Setup.}
\label{app:experiment_setup}
\textbf{Task-Specific Details.}
We adopt different strategies based on the characteristics of each task. 
For Data Comprehension and Code Capability, we employ prompt-based evaluation methods as these tasks can be effectively assessed without additional training. 
\underline{In the case of \textit{Data Comprehension}}, we observe that when provided with file data, the model often redundantly attempts to reload the table using Python, leading to context overflow and reduced task accuracy. 
Moreover, fine-tuning may not reliably enhance the model's ability to comprehend data and may inadvertently reinforce these redundant behaviors, making prompt-based evaluation a more suitable choice.
\underline{For \textit{Code Capability}}, the current selection of models already supports effective evaluation of coding abilities, making fine-tuning unnecessary. 
\underline{For \textit{Strategic Planning}}, we use LoRA~\cite{lora} fine-tuning since this task requires the model to learn complex reasoning patterns and long-term dependencies through targeted training. 
To ensure fair comparisons across experiments, dataset sizes are controlled to be equal within the same task setting.
However, some unavoidable differences exist across tasks due to varying data requirements and availability.

\begin{table}[!ht]
\centering
\def\arraystretch{1.0}%
\scalebox{1}{ 
\begin{tabular}{l r r}
\toprule
Source & \# Question & \# Sample \\
\midrule
DAEval &  257 &  228 \\
DSBench &  466 &  319 \\
TableBench &  886 &  612 \\
WTQ &  4,344 &  3,779 \\
FetaQA &  2,000 &  1,505 \\ \midrule
Total & 7,899 &  6,443 \\
\bottomrule
\end{tabular}
}
\caption{Statistics of the generated data size.}
\label{tab:data_stat} 
\end{table}

\textbf{Datasets Collection.}
\label{app:data_collect}
We synthesized a series of trajectories based on existing data. 
Notably, we exclude portions of DSBench data that involve image inputs to maintain consistency and ensure the focus of our analysis.
Detailed statistics of the generated data can be found in Table~\ref{tab:data_stat}.

\textbf{Training and Evaluation Details.}
For fine-tuning, we employ the LLaMA Factory framework~\citep{zheng2024llamafactory}. 
Experiments are conducted on 4 NVIDIA A800 80G GPUs, leveraging the DeepSpeed ZeRO-3 optimization for efficient distributed training. We adopt a per-device batch size of 1, with gradient accumulation steps set to 4. 
Training is performed using bfloat16 precision, and a warmup ratio of 0.02. 
We employ a cosine learning rate scheduler with an initial learning rate of $1 \times 10^{-5}$.
All our fine-tuned models are evaluated using vLLM~\cite{vllm} with the temperature set to 0.

\section{Problem Diversity.}
\label{app:problem_diverse}

To quantify diversity, we employed a three-stage procedure to classify each question into distinct domains. 
In the first stage, we asked the model to classify a small sample of questions without any constraints; 
Based on its raw outputs, we manually merged semantically similar concepts and split those that were overly broad, yielding a preliminary set of reference categories. 
In the second stage, we iteratively refined this taxonomy: the model classified each question according to the current reference set but was permitted to propose new categories for outliers, which we then reviewed and either incorporated into or merged with the existing categories. 
This cycle continued until no further category labels emerged. 
In the third stage, we constrained the model to assign every question exclusively to one of the finalized categories. 
Applying this pipeline to our full dataset yielded 10 distinct domains as shown in Figure~\ref{categories}.

\begin{figure*}[!htbp]
\centering
\scalebox{1}{
\begin{tcolorbox}
Here we present the specific descriptions of each category.

\textbf{Data Profiling}: Systematically examining dataset characteristics (e.g., completeness, distributions, anomalies) to understand its fundamental structure.  

\textbf{Data Retrieval}: Extracting specific data subsets from storage systems using query-based methods.  

\textbf{Data Aggregation}: Combining data from multiple sources into unified formats for analytical purposes.  

\textbf{Causal Analysis}: Identifying cause-effect relationships between variables through controlled experiments or counterfactual reasoning.  

\textbf{Exploratory Analysis}: Preliminary investigation using visualization and summary statistics to formulate hypotheses.  

\textbf{Predictive Analysis}: Building models to forecast future outcomes based on historical patterns.  

\textbf{Inferential Analysis}: Drawing population-level conclusions from sample data via statistical hypothesis testing.  

\textbf{Variance Analysis}: Quantifying and attributing variability in data to specific influencing factors.  

\textbf{Pattern Description}: Detecting and formally characterizing recurring structures/relationships in data.  

\textbf{Data Visualization}: Creating graphical representations to communicate complex information efficiently. 

\end{tcolorbox}
}

\caption{Data Analysis Task Categories.}
\label{categories}
\end{figure*}


\section{Data Synthesis Details}
\label{app:data_sft}

\begin{algorithm}[!h]
\caption{Data Generation Pipeline}
\label{alg:response_pipeline}
\small
\textbf{Input}: Question set $\mathcal{Q}$; File set $\mathcal{F}$ \\
\textbf{Output}: Final responses $\mathcal{R}^{*}$ (each with formatted reasoning summary and code segments)
\begin{algorithmic}[1]
\STATE Initialize empty output set $\mathcal{R}^{*} \gets \emptyset$
\FOR{each question $q \in \mathcal{Q}$}
    \STATE $R_q \gets \text{GenerateResponses}(q, \mathcal{F})$
    \STATE $R_q \gets \text{FilterCorrectResponses}(R_q)$
    \IF{$R_q = \emptyset$}
        \STATE \textbf{continue}
    \ENDIF
    \FOR{each response $r \in R_q$}
        \STATE $r_{\text{clean}} \gets \text{ApplyRulesAndFormat}(r)$
        \IF{$\text{IsLowComplexity}(r_{\text{clean}})$ or $\text{NotMediumTurn}(r_{\text{clean}})$}
            \STATE \textbf{continue}
        \ENDIF
        \STATE $reason_{\text{chain}} \gets \text{ExtractReasonChain}(r_{\text{clean}})$
        \STATE $reason_{\text{summary}} \gets \text{Summarize}(reason_{\text{chain}})$
        \STATE $r_{\text{final}} \gets \text{InsertBeforeCode}(r_{\text{clean}}, reason_{\text{summary}})$
        \STATE Add $r_{\text{final}}$ to $\mathcal{R}^{*}$
    \ENDFOR
\ENDFOR
\STATE \textbf{return} $\mathcal{R}^{*}$
\end{algorithmic}
\end{algorithm}

Our data synthesis pipeline (see Algorithm~\ref{alg:response_pipeline}) transforms raw queries into high-quality instances for Supervised Fine-Tuning. The process consists of three core stages:

\begin{enumerate}
    \item \textbf{Prompt-Based Answer Generation:} For each input question $q \in \mathcal{Q}$, we use a \textbf{base generation prompt} (Figure~\ref{prompt:base}) and a non-zero temperature (e.g., $T=0.6$) to generate a set of diverse candidate responses $R_q = \{r_1, \dots, r_k\}$, where we set $k=3$ in our experiments.

    \item \textbf{Targeted Instance Selection:} Next, we conduct a multi-faceted filtering process where each candidate response must pass the following sequential criteria to be selected:
    \begin{itemize}
        \item \textbf{Correctness Verification:} First, as a crucial preliminary step, we execute the code within each candidate response to verify its correctness against a ground-truth solution. Only demonstrably correct responses proceed to the subsequent filters.
        
        \item \textbf{Format Conformance:} Next, we apply a set of rules to each verified response to standardize its structure, producing a cleaned version, $r_{\text{clean}}$. Responses that fail to conform to our required format (e.g., missing or malformed code blocks) are discarded.
        
        \item \textbf{Length and Difficulty Filtering:} Finally, the cleaned response $r_{\text{clean}}$ is subjected to two concurrent checks. We retain only those spanning 4--5 turns (medium length) and discard any solvable by a less capable baseline model (Qwen2.5-7B-Instruct) to ensure medium-to-high difficulty.
    \end{itemize}
    Only the responses that pass all these filters are passed to the next stage.
    \item \textbf{Reasoning-Driven Data Enrichment:} For each filtered response $r_{\text{clean}}$, we first extract its detailed reasoning (\texttt{reason\_chain}) and then condense it into a concise summary, \texttt{reason\_summary}. This is accomplished using a dedicated \textbf{summarization prompt} (Figure~\ref{prompt:summaization}). The final instance, $r_{\text{final}}$, is formed by inserting this summary into the response, explicitly capturing the underlying problem-solving strategy.
\end{enumerate}

This pipeline yields a final dataset of 2.8k instances used for our fine-tuning experiments.

\begin{table}[!h]
    \centering
    \scalebox{.9}{
    \begin{tabular}{c|c|c}
        \toprule
        \textbf{Hyperparameter} & \textbf{7B} & \textbf{14B} \\
        \midrule
        batch size & 32 & 32 \\
        batch size per device & 1 & 1 \\
        gradient accumulation steps & 8 & 4 \\
        learning rate & \(5.0 \times 10^{-6}\) & \(1.0 \times 10^{-5}\) \\
        epochs & 3 & 3 \\
        warmup ratio & 0.1 & 0.1 \\
        bf16 & \texttt{true} & \texttt{true} \\
        \bottomrule
    \end{tabular}
    }
    \caption{Detailed training hyperparameters.}
    \label{tab:hyperparameters}
\end{table}

\paragraph{Training Details.} For model fine-tuning, the 7B model was trained using 4 NVIDIA A800 80G GPUs, while the 14B model utilized 8 NVIDIA A800 80G GPUs. The training configurations are summarized in Table~\ref{tab:hyperparameters}.

\begin{figure*}[!h]
\centering
\scalebox{1}{
\begin{tcolorbox}

You are an expert data analyst tasked with solving complex analytical challenges. Your approach requires careful data investigation, rigorous statistical analysis, thorough validation, evidence-based decision making, and clear documentation.

**Response Structure**:

1. Begin each step with "\#\# Thought: " followed by **a thoughtful narrative explanation** of your reasoning and approach, using natural language to walk through your thought process. 

\quad - Clearly explain your current thinking and analytical approach

\quad - Include why you chose this approach and what alternatives you considered  
   
\quad - Explain how this step connects to your overall analysis strategy
   
\quad - Present your thinking as a fluid, well-structured narrative

2. Follow this explanation with the corresponding Python code in \textquotesingle{}\textquotesingle{}\textquotesingle{}python \textquotesingle{}\textquotesingle{}\textquotesingle{} tags:

\quad - Include code ONLY when it contributes to solving the task
   
\quad - Every code block MUST include print() statements with clear labels
   
\quad - Ensure code is focused and uses accurate syntax
     
3. Await "\#\# Observation:" sections to review outputs before continuing analysis.

4. Conclude your response with "\#\# Final Answer:", which should:

\quad - Include a brief summary of the analysis process

\quad - Provide the final, precise solution backed by evidence

Format example:

\#\# Thought: [Description]

\#\# Code: 

\textquotesingle{}\textquotesingle{}\textquotesingle{} \  python

[code if needed]

\textquotesingle{}\textquotesingle{}\textquotesingle{} \

\#\# Thought: [Description]

\#\# Code: 

\textquotesingle{}\textquotesingle{}\textquotesingle{} \ python

[code if needed]

\textquotesingle{}\textquotesingle{}\textquotesingle{} \ 

...

\#\# Final Answer: [Your final answer]

\end{tcolorbox}
}
\caption{Prompt for Data Generation.}
\label{prompt:base}
\end{figure*}

\begin{figure*}[!h]
\centering
\scalebox{1}{
\begin{tcolorbox}

Please reconstruct the given reasoning content according to the following two steps:

1. Extract the main process and key steps from the reasoning content

2. Reconstruct the complete reasoning process based on these key steps
~\\

<Requirements>

Express the reconstructed reasoning process in a more concise way, requiring:

\quad - Retain all effective reasoning information

\quad - Ensure logical coherence and consistency

\quad - Maintain the same meaning as the original reasoning content

\quad - Remove redundant and repetitive content

\quad - Present the entire reasoning process with a clear structure in a few short sentences

\quad - Keep the tone of the original thought process

</Requirements>
~\\

<Output\_format>

\#\# Reconstruction: [Only Final Reconstruction Results]

</Output\_format>
~\\

Here is the reasoning content:

\{reasoning\_content\}


\end{tcolorbox}
}

\caption{Prompt for Summarization.}
\label{prompt:summaization}
\end{figure*}

\end{document}